\documentclass[letterpaper, 10 pt, conference]{ieeeconf}  
\IEEEoverridecommandlockouts                              
\usepackage{multirow}
\usepackage[colorlinks=true, bookmarks=true]{hyperref}

\usepackage{color,xcolor}
\usepackage{epsfig}
\usepackage{graphicx}

\usepackage{adjustbox}
\usepackage{array}
\usepackage{booktabs}
\usepackage{colortbl}
\usepackage{float,wrapfig}
\usepackage{hhline}
\usepackage{multirow}
\usepackage{subcaption} 
\usepackage[font={small}]{caption}

\usepackage{amsmath,amsfonts,amssymb}
\usepackage{bm}
\usepackage{nicefrac}
\usepackage{microtype}

\usepackage{changepage}
\usepackage{extramarks}
\usepackage{fancyhdr}
\usepackage{lastpage}
\usepackage{setspace}
\usepackage{soul}
\usepackage{xspace}

\usepackage{url}

\usepackage{algorithm, algorithmic}
\usepackage{enumerate}
\usepackage{todonotes} 
\usepackage{gensymb}
\usepackage{color,soul}

\newcolumntype{L}[1]{>{\raggedright\let\newline\\\arraybackslash\hspace{0pt}}m{#1}}
\newcolumntype{C}[1]{>{\centering\let\newline\\\arraybackslash\hspace{0pt}}m{#1}}
\newcolumntype{R}[1]{>{\raggedleft\let\newline\\\arraybackslash\hspace{0pt}}m{#1}}

\newcommand{\ignorethis}[1]{}

\makeatletter
\DeclareRobustCommand\onedot{\futurelet\@let@token\@onedot}
\def\@onedot{\ifx\@let@token.\else.\null\fi\xspace}

\makeatother

\definecolor{MyDarkBlue}{rgb}{0,0.08,1}
\definecolor{MyDarkGreen}{rgb}{0.02,0.6,0.02}
\definecolor{MyDarkRed}{rgb}{0.8,0.02,0.02}
\definecolor{MyDarkOrange}{rgb}{0.40,0.2,0.02}
\definecolor{MyPurple}{RGB}{111,0,255}
\definecolor{MyRed}{rgb}{1.0,0.0,0.0}
\definecolor{MyGold}{rgb}{0.75,0.6,0.12}
\definecolor{MyDarkgray}{rgb}{0.66, 0.66, 0.66}

\title{\LARGE \bf
SwingBot: Learning Physical Features from In-hand Tactile Exploration for Dynamic Swing-up Manipulation}

\author{Chen Wang$^{*1,2}$, Shaoxiong Wang$^{*1}$, Branden Romero$^{1}$, Filipe Veiga$^{1}$ and Edward Adelson$^{1}$ \\
\href{http://gelsight.csail.mit.edu/swingbot}{http://gelsight.csail.mit.edu/swingbot}
\thanks{$^{*}$Authors with equal contribution.}
\thanks{$^{1}$CSAIL, Massachusetts Institute of Technology}%
\thanks{$^{2}$Department of Computer Science, Shanghai Jiao Tong University}%
}

\begin{document}

 \maketitle
\thispagestyle{empty}
\pagestyle{empty}


\begin{abstract}

Several robot manipulation tasks are extremely sensitive to variations of the physical properties of the manipulated objects. One such task is manipulating objects by using gravity or arm accelerations, increasing the importance of mass, center of mass, and friction information. We present SwingBot, a robot that is able to learn the physical features of a held object through tactile exploration. Two exploration actions (tilting and shaking) provide the tactile information used to create a physical feature embedding space. With this embedding, SwingBot is able to predict the swing angle achieved by a robot performing dynamic swing-up manipulations on a previously unseen object. Using these predictions, it is able to search for the optimal control parameters for a desired swing-up angle. We show that with the learned physical features our end-to-end self-supervised learning pipeline is able to substantially improve the accuracy of swinging up unseen objects. We also show that objects with similar dynamics are closer to each other on the embedding space and that the embedding can be disentangled into values of specific physical properties.

\end{abstract}

\section{Introduction}
\label{sec:intro}

As applications for robotic manipulation shift from industrial to service tasks, the need for robots to deduce the physical properties of objects increases. To cope with the diversity of objects and tasks in the real world, robots require models that can quickly infer the physical properties of objects, with as few interactions as possible and without explicit supervision. These models could allow the robot to perform more dynamic interactions with its environment or with held objects in the cases where in-hand manipulation is desired. Vision based methods for learning physical object representations through dynamic interaction have shown some promise towards achieving such models~\cite{xu2019densephysnet}. However, vision based approaches are still restricted to interactions in structured environments and do not address the limitations of deploying deep learning based vision systems into the real-world scenarios. 

Tactile sensing can be seen as an attractive alternative to vision. In particular, vision-based tactile sensors provide direct observations of the deformation caused by contact with an object~\cite{yuan2017gelsight}. Considering the local nature of these observations, the influence of environmental noise is negligible, making methods developed with this modality potentially more transferable to real-world environments. Additionally, vision-based tactile sensors are able to accurately estimate the normal and shear forces being applied to the sensing surface. So rather than designing an environment to make the influences of external forces easily observable with vision, it is preferable to have very accurate sensing directly at the interaction points, i.e, performing interactions with a sensorized hand. Therefore, tactile sensing seems like an appropriate modality for learning object physical representations. However, it is not without its limitations as these sensors are soft, making the modeling and measuring of the properties of the sensor itself more complex.

\begin{figure}[t]
	\centering
	\includegraphics[width=1.0\linewidth]{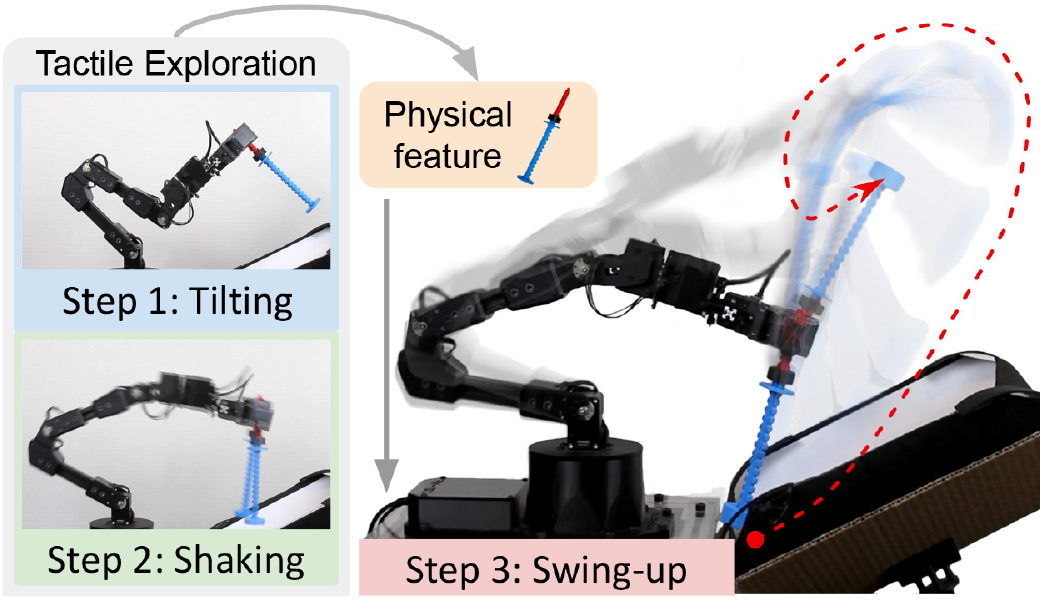}
    \vspace{-13pt}
	\caption{\textbf{SwingBot.} We develop a learning-based in-hand physical feature exploration method with a GelSight tactile sensor, which assists the robot to perform accurate dynamic swing-up manipulation.}
	\vspace{-10pt}
	\label{fig:pullfig}
\end{figure}

\def\svgwidth{\linewidth}
\begin{figure*}[t]
	\centering
\begingroup%
  \makeatletter%
  \providecommand\color[2][]{%
    \errmessage{(Inkscape) Color is used for the text in Inkscape, but the package 'color.sty' is not loaded}%
    \renewcommand\color[2][]{}%
  }%
  \providecommand\transparent[1]{%
    \errmessage{(Inkscape) Transparency is used (non-zero) for the text in Inkscape, but the package 'transparent.sty' is not loaded}%
    \renewcommand\transparent[1]{}%
  }%
  \providecommand\rotatebox[2]{#2}%
  \newcommand*\fsize{\dimexpr\f@size pt\relax}%
  \newcommand*\lineheight[1]{\fontsize{\fsize}{#1\fsize}\selectfont}%
  \ifx\svgwidth\undefined%
    \setlength{\unitlength}{694.51551819bp}%
    \ifx\svgscale\undefined%
      \relax%
    \else%
      \setlength{\unitlength}{\unitlength * \real{\svgscale}}%
    \fi%
  \else%
    \setlength{\unitlength}{\svgwidth}%
  \fi%
  \global\let\svgwidth\undefined%
  \global\let\svgscale\undefined%
  \makeatother%
  \begin{picture}(1,0.26737682)%
    \lineheight{1}%
    \setlength\tabcolsep{0pt}%
    \put(0,0){\includegraphics[width=\unitlength,page=1]{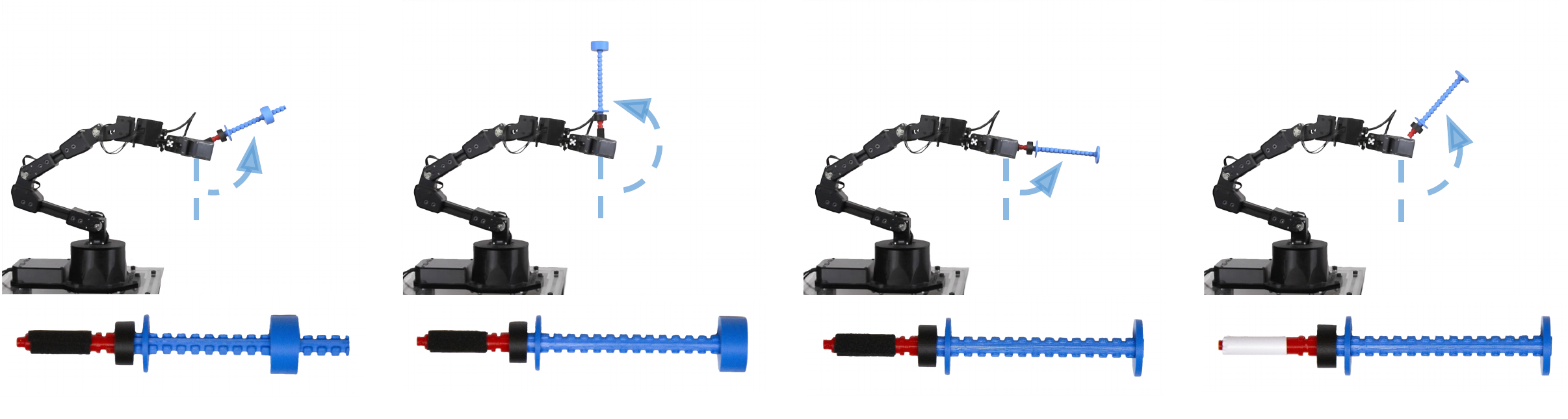}}%
    \put(0.10200162,0.00363942){\color[rgb]{0,0,0}\makebox(0,0)[lt]{\smash{\begin{tabular}[t]{l}\scriptsize (a)\end{tabular}}}}%
    \put(0.35732425,0.00362887){\color[rgb]{0,0,0}\makebox(0,0)[lt]{\smash{\begin{tabular}[t]{l}\scriptsize (b)\end{tabular}}}}%
    \put(0.61514483,0.00363942){\color[rgb]{0,0,0}\makebox(0,0)[lt]{\smash{\begin{tabular}[t]{l}\scriptsize (c)\end{tabular}}}}%
    \put(0.86896025,0.00362887){\color[rgb]{0,0,0}\makebox(0,0)[lt]{\smash{\begin{tabular}[t]{l}\scriptsize (d)\end{tabular}}}}%
  \end{picture}%
\endgroup%

	\caption{\textbf{Challenges.} Swing-up is a highly-dynamic process, where changing objects' physical properties would have a big impact on the final swing-up angle. Here we show, with the same control parameters, that the dynamics vary when the objects vary: same mass but different center of mass (a)(b); different mass (b)(c); and different friction coefficient (c)(d).}
	\vspace{-10pt}
	\label{fig:challenges}
\end{figure*}
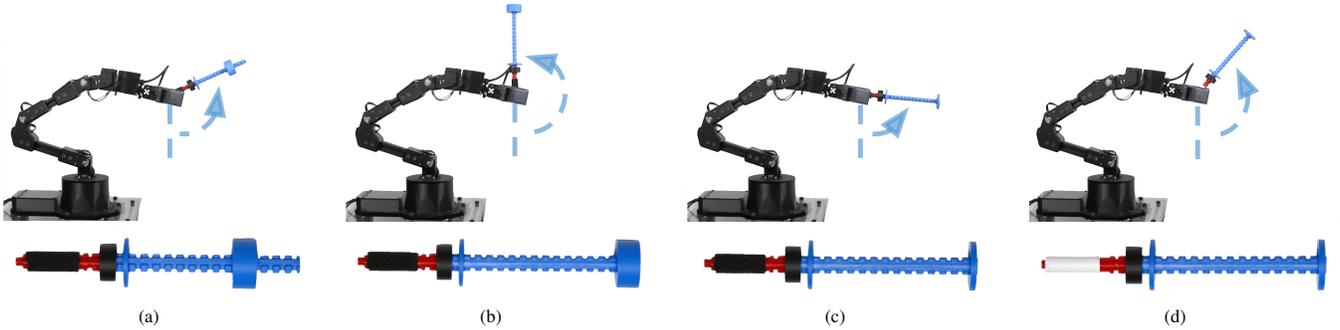
In this work we develop a method to infer the physical parameters of an unknown object through in-hand exploration. To do this, we use the information provided by a GelSight sensor~\cite{yuan2017gelsight} to learn a low-dimensional embedding of the object's properties as well as the properties of the GelSight itself. We learn the embedding in a self-supervised fashion and use it to optimize the performance of a dynamic in-hand manipulation task. In particular we have the robot swing-up a set of unknown objects to a desired pose in-hand. We find the optimal control parameters for the swing-up task with the aid of a swing-up angle predictor that uses our learned embedding as input. We also prove the portability of this embedding to new tasks by showing that we can use it to directly regress to object parameters such as mass, center of mass, moment of inertia and friction. 

Our approach consists of two main components: (1) an information fusion model and (2) a forward dynamics model. SwingBot starts by performing two in-hand exploration actions, tilting and shaking. As each of these actions provides different information about the physical parameters of the object, a fusion model takes the information from both actions in order to learn a joint physical feature embedding of the object in-hand. Once the embedding is learned, a forward dynamics model uses the embedding and the control parameters that generate the swing-up motion in order to predict the final swing-up angle.

The main contribution of this work is to demonstrate that the robot is able to learn a low-dimensional embedding of the physical features of a held object from dense tactile feedback acquired through a small number of active exploration actions. The learned embedding allows the robot to accurately and consistently perform a swing-up task on a set of objects, with an overall 17.2 degree error on unseen objects. Furthermore, our experiments show that the fusion network can accurately estimate physical parameters of unknown objects once the features are disentangled.

\section{Related Work}
\label{sec:related}

Robotic manipulation has been dominated by the paradigm of kinematic manipulation, and for good reasons. In reducing the effects of task dynamics, it is easier to ensure that a robot can perform its task consistently without error. However, this has limited the application of robotics to a set of simple tasks like pick-and-place. As robotic manipulation becomes more ubiquitous, the need for robots that can perform more tasks becomes important. One path forward is to increase the mechanical complexity of robots by using dexterous hands. However this also comes with a cost in terms of control and design complexity. Alternatively,~\cite{mason1993dynamic} illustrates that simple mechanical designs can achieve more than pick-and-place if we reconsider the task dynamics. 

Inspired by~\cite{mason1993dynamic}, researchers have been successful in developing methods that exploit the task dynamics for performing actions like dynamically sliding an object in-hand~\cite{shi2017dynamic}, tossing an object into the air to regrasp it~\cite{dafle2014extrinsic}, and swinging up an object to a desired pose~\cite{sintov2016robotic}. However, these methods require experts to first determine which parameters of the system are important for the task, a model of the dynamics, and accurate measures of the physical properties of importance for each object used. Therefore, these methods are hard to deploy in real-world environments. 

To alleviate the need for careful modelling and accurate measurements, researchers have been working on an alternative method known as intuitive physics~\cite{wu2015galileo, agrawal2016learning}. Intuitive physics allows a robot to estimate the parameters of an object via learning based approaches and interaction. In~\cite{wu2015galileo, lerer2016learning, xu2019densephysnet, fragkiadaki2015learning},  direct regression over the physical parameters of an object, like mass and friction, was performed for tasks like sliding an object and predicting the stability of a tower of stacking blocks. However, knowing exactly which physical parameters are needed for a task or directly observing those parameters from feedback may be difficult. So, several methods~\cite{agrawal2016learning, finn2017deep, yen2019experience,zeng2019tossingbot} instead indirectly estimate object parameters by learning an object embedding in a self-supervised way for tasks like pushing and tossing an object to a desired pose. However, these methods still require a structured environment. In particular,~\cite{xu2019densephysnet} used a set of ramps to make the result of a dynamic interaction easily observable with vision. 

Rather then using the environment we can instead use in-hand manipulation to extract properties about the object. In fact,~\cite{lederman1987hand} suggest that humans perform a set of exploratory procedures to extract object properties like friction, mass and center of mass. While, it is possible to monitor in-hand interactions with vision,~\cite{calandra2018more} shows tactile sensing outperforms vision alone when doing tasks that require feedback about contact interactions like determining if a grasp is successful. This work along with other works that explore tactile sensing for tasks like slip control~\cite{veiga2018grip, dong2019maintaining, stepputtis2018extrinsic}, regrasping~\cite{chebotar2016self, dang2013grasp, hogan2018tactile}, contour following~\cite{lepora2019pixels, hellman2017functional, she2019cable}, and ball manipulation~\cite{tian2019manipulation} focus mainly on static or quasi-static interactions. The object's physical properties have less of an influence on the performance of a controller for static or quasi-static interactions then they would have in more dynamic manipulation tasks like swing up. As a result, none of aforementioned works that explore tactile sensing estimate the physical parameters of the object. In contrast, we focus on learning physical representations from simple in-hand tactile exploration, and show that such representations are useful for manipulation tasks that requires physical knowledge.

\section{Method}
\label{sec:method}

\def\svgwidth{\linewidth}

\begin{figure*}[t]
	\centering
	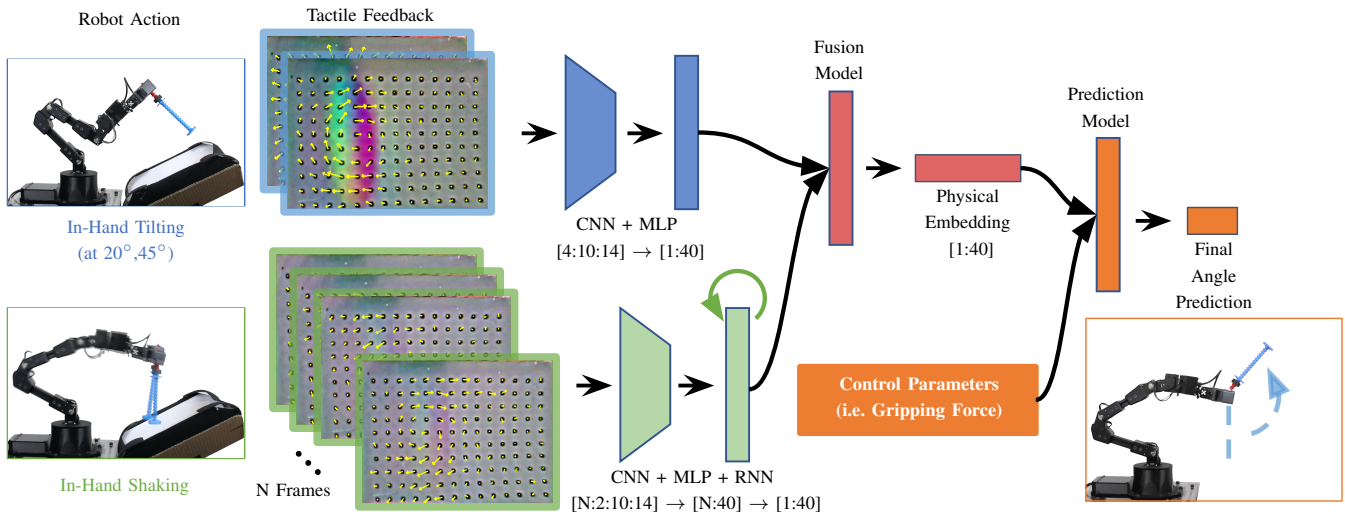
	\caption{\textbf{Overview of the architecture.} The robot takes several steps to acquire and use the physical features of the held object: (1) \textit{Tilting} the object at 20\degree and 45\degree. The corresponding marker information is encoded by a network with CNN and MLP into a 40-dimensional embedding. (2) \textit{Shaking} the object. The sequence of marker information is processed by a RNN network into a 40-dimensional embedding. (3) A fusion model concatenates the embedding from both actions and outputs a fused physical feature embedding. (4) A prediction model takes the physical embedding and control paramaters as input and outputs a prediction of the final swing-up angle. \textbf{During training}, the whole pipeline is trained in an end-to-end fashion using the final angle for self-supervision. \textbf{During inference}, a set of control parameters are uniformly sampled. The action with the prediction result closest to the goal is selected to perform the swing-up.}
	\vspace{-10pt}
	\label{fig:model}
\end{figure*}

\def\svgwidth{\linewidth}

The goal of SwingBot is to enable the robot to swing up an unknown object to a desired pose ($0\degree \sim 200\degree$) after performing a single exploratory action. In~\cite{sintov2016robotic}, the authors suggest the robot must first build a dynamic model of the task, and once the robot has a notion of what this model is, it must then extract which physical parameters of an object are keys to completing the task. Thus, when a novel object is introduced, the system only needs to extract those parameters to tune the model. Therefore, we create a method to estimate the desired control parameters of a hand coded control policy by performing a set of hand coded exploration actions. To accomplish this we use GelSight, a vision based tactile sensor, to monitor the state of the object while performing in-hand exploration of the object in the form of shaking and tilting. These exploratory procedures extract different type of object information, and as a result we create a method to fuse the information from both procedures into a joint physical feature embedding of the object. We then create a forward dynamics model that uses the embedding to infer which action will result in our desired object pose.


\subsection{GelSight}
\begin{figure}[t]
	\centering
	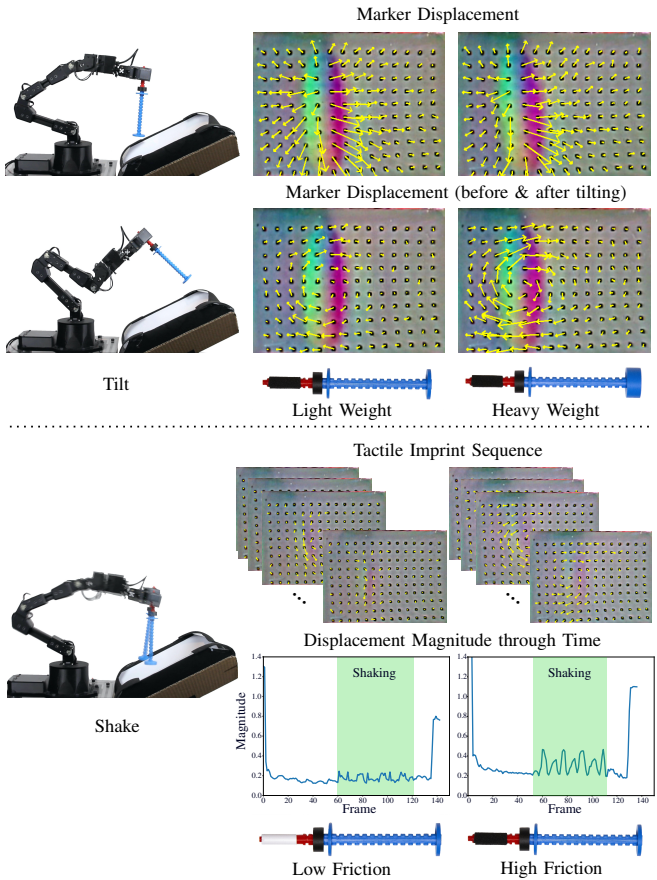
    \vspace{-13pt}
	\caption{\textbf{Exploration actions and GelSight Signals.} The robot executes two in-hand explorations, tilting and shaking, to acquire tactile observations of the object. When tilting, different force and torque distributions are generated by the objects weight can be observed. When shaking, different frictions and vibrations can be observed from temporal sequences of tactile signals.}
	\vspace{-10pt}
	\label{fig:exploration}
\end{figure}

While previous methods exploring physical object property estimation monitored the result of a dynamic interaction with vision \cite{xu2019densephysnet}, vision as a modality has its limitations for this task. Beyond errors in state estimation due to environmental noise, it lacks the ability to perceive the forces being applied to an object. Hence, if you were performing an exploratory action like tilting an object in-hand to estimate it's mass, its change in position as you tilt the object would be almost imperceptible, as seen in Fig~\ref{fig:exploration}. Therefore, we rely on tactile sensing, the GelSight sensor~\cite{yuan2017gelsight} in particular. The GelSight enables us to have high resolution information about the contact surface between the object and the finger. This enables us to have information about local geometry of the object for pose estimation. Beyond that the sensor used in this experiment is equipped with markers along the sensing surface which provides information about tangential displacements, giving us rich information about the sheer forces and torques being applied to the sensing surface. 

\subsection{Information Fusion for Multiple Exploration Actions}

While the use of a GelSight has its advantages in providing rich information about the contact dynamics between the finger and the object, it also comes with its limitations. The material used to make the GelSight (Polydimethylsiloxane) exhibit nonlinear mechanical properties that are difficult to measure and model. So, while previous approaches~\cite{xu2019densephysnet} were able to directly regress over physical properties like mass and friction and perform a forward simulation, we take a different approach. Rather than regressing over the physical parameters, we hand design a set of exploratory procedures that clearly encode physical properties of the object like fiction and mass, and then let the model create its own low-dimensional embedding of the object using self-supervised learning in hopes it also encodes the gel's dynamics.

In designing these exploration actions, we had to determine what set of parameters to search for. In \cite{sintov2016robotic}, a dynamic analysis of the swing-up task was performed, concluding that the \textbf{surface friction}, \textbf{mass of the object}, \textbf{center of mass} and \textbf{moment of inertia} play roles in swing-up dynamics. Since we use a parallel gripper for this task we are limited to what we can choose in terms of actions. We determined that shaking and tilting the object in hand were the only methods that can be performed reliably. After some experimentation with these behaviors, we determined that tilting was able to give us information about \textbf{mass}, \textbf{center of mass} and the \textbf{moment of inertia}, while shaking is able to inform us of the \textbf{friction} of the object as show in Fig.~\ref{fig:exploration}.


\begin{figure}[t]
	\centering

	\includegraphics[width=1.0\linewidth]{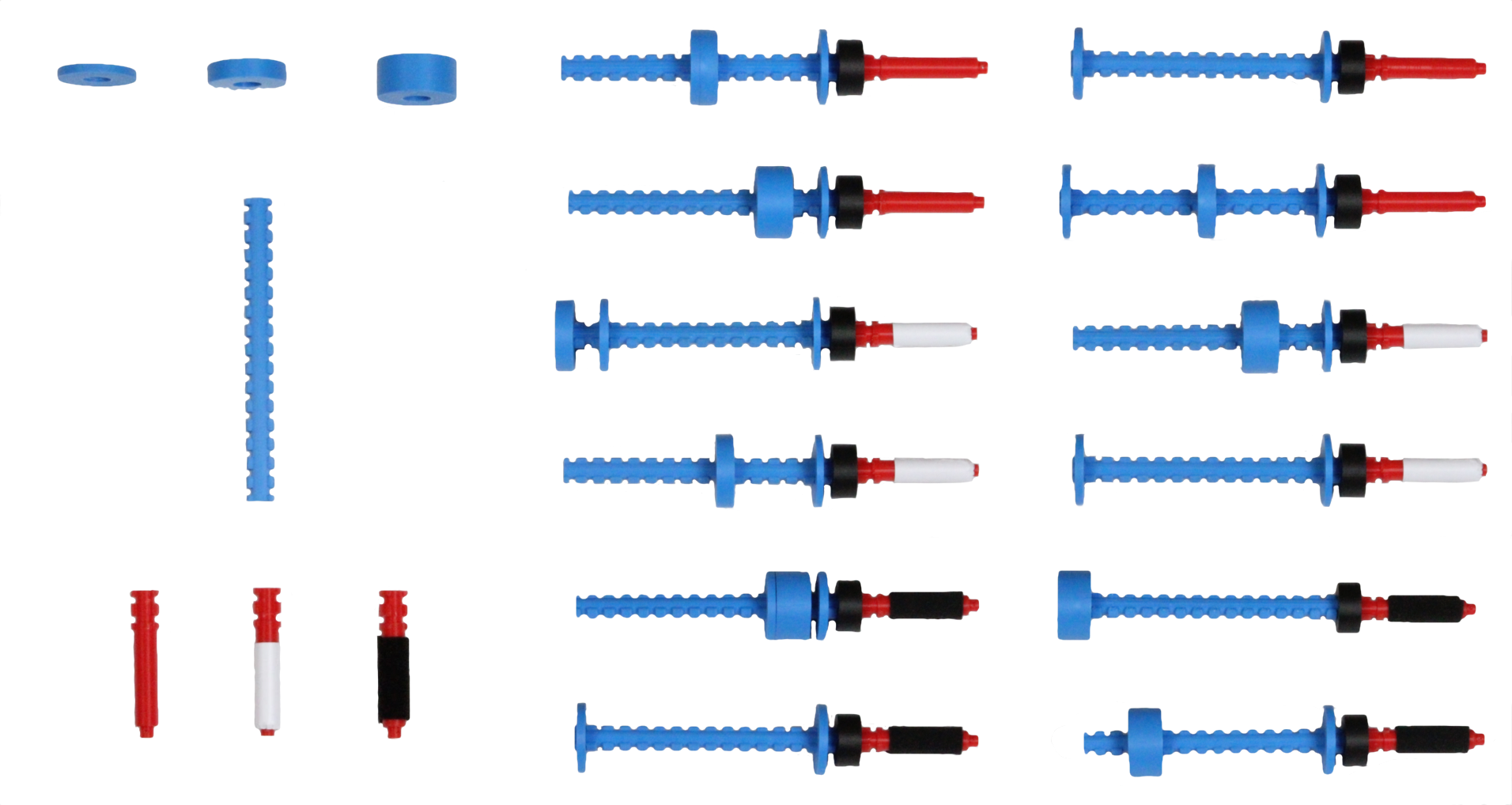}
    \vspace{-13pt}
	\caption{\textbf{Template objects.} The template objects consist of three components: handle, rack and weights. Different components can be assembled and replaced easily, which creates a variety of objects with different physical properties.}
	\vspace{-10pt}
	\label{fig:dataset}
\end{figure}
\noindent\textbf{In-hand Object Tilting:} Using tactile feedback and tilting the object in-hand to different angles provides us information about the \textbf{mass} and \textbf{center of mass}, as showed in Fig.~\ref{fig:exploration}. We observed that as the object was tilted to a low angle we could obtain its mass, while tilting the object to a larger angle gave us information about the torque being applied to the sensor. Combining mass in torque estimates we were then able to infer the center of mass. Therefore, after the robot grasped the object and held it in-hand, it is then able to tilt the object into 20 and 45 degree poses. The marker feedback ($W \times H \times 2$; $W=14, H=12$ in our experiments) from the GelSight tactile sensor at each angle is recorded and used as the input information to the model. Then, the model concatenates the marker information of three angles into a 4 dimensional inputs, followed by a CNN network with kernel size of $5 \times 5$, $3 \times 3$ and $2 \times 2$. The last layer of the network is a fully-connected layer which outputs a 40-dimensional feature as the fusion of the learned physical proprieties for tilting.

\noindent\textbf{In-hand Object Shaking:} Shaking in turn contains information about the \textbf{friction}, and potentially for the \textbf{moment of inertia}. After holding the object in a 0-degree pose, the robot first loosens the gripper force to enable a small range of rotation flexibility (Fig.~\ref{fig:exploration}). Then, the robot starts a quick switch between forward and backward rotations (5 degrees in our experiments) on the joint of the end effector. During this process, we record a sequence of the tactile marker displacements (60-70 frames per trial). Each frame is then processed into a 40-dimensional embedding with the CNN network, which has the same architecture as the one introduced above. Since we want to extract the inter-frame information of the shaking action, we use a long-short term memory (LSTM)~\cite{hochreiter1997long}, which starts with zero hidden states and iteratively processes all the embeddings of each frame. The last layer will concatenate the hidden states $h$ and cell state $c$ into a 80-dimensional embedding as the fusion of the learned physical properties for shaking.

\subsection{Prediction Model for Forward Dynamics}
To perform the swing-up action we use an impulse-momentum method~\cite{albahkali2009swing}. The first stage of the swing-up action begins by having the robot build up the object's linear and rotation momentum by simultaneously accelerating the object upwards and rotating the wrist in the direction of the swing while holding the object firmly. After a short period the robot creates an impulse, by quickly accelerating the object downwards and rotating the wrist in the opposite direction of the swing. At the moment of the impulse the robot loosens the gripper, so that the inertia of the object can overcome the forces of rotational friction and gravity. Thus, the object freely rotates in-hand. After some time, the gripper is tightened to stop the motion of the object at some pose. We use current based position control for the gripper so that the robot automatically decides the gripper width for holding different objects tightly with the same motor torque. When designing the action, the linear and rotational movements of the arm are predefined as well as the timing of gripper tightening, but the robot selects how much the gripper loosens at the impulse. This allows the robot to control the objects deceleration so that it can precisely control the object's end pose.

In order to use the learned physical features to find the control parameter, the gripper width, for the swing-up manipulation, we propose a forward dynamics model that takes the fused physical features and the action as inputs and outputs a prediction of the final swing-up angle (Fig.~\ref{fig:model}). The model is trained in a self-supervised learning fashion. The data collection is introduced in Sec.~\ref{sec:experiments}. During the inference mode, the robot first records the marker information of the tilting and shaking actions. Then, the trained information fusion model processes these inputs into a joint physical feature. After that, a set of gripper widths are uniformly sampled. The prediction model predicts the final swing-up angle for each sampled gripper width and then selects the one with the prediction result closest to the goal pose.

\subsection{Template Objects and Dataset}  
When it comes to model generalization, the diversity of the training conditions highly influences the models performance on unseen objects. To this end, inspired by~\cite{bauza2018omnipush}, we design a modular system to quickly build a set of test objects. Our object templates are shown in Fig~\ref{fig:dataset}. There are three major components: handle, rack, and weight. They can be assembled or replaced by simple rotational press-fit. The goal is to change the object's physical properties easily by placing different weights in different positions and exchanging handles.

With our template objects, we collect a dataset that contains 33 different objects and each object was used in 50 swing-up trials, performed with a random control parameter. These objects contain variance in different category of physical proprieties:

\begin{itemize}
    \item 3 different \textbf{surface frictions} on the handle: foam, slick tape, and plastic.
    \item 3 disks with different \textbf{mass}: 3.7~g, 7.3~g and 14.5~g. 
    \item a pole-shaped rack (15.6~g) allowing for different placement of the disks for variance in \textbf{center of mass} (77-134 mm) and \textbf{moment of inertia} (0.03-0.58 $g\cdot m^2$)
\end{itemize}

In each data collection trial, the robot first grasps the object and holds it a 0-degree pose. It then rotates its end effector into two angles (20\degree, 45\degree in our experiments), as introduced in Sec.~\ref{sec:method}, and records the marker information from the tactile sensor. After that, the robot resets the object pose to 0\degree and loosens the gripper force before it starts shaking as introduced in Sec.~\ref{sec:method}. The marker sequence is recorded. Then the robot selects a random control parameter and starts its swing up. The final angle in the end of the swing-up is saved as the supervision ground truth for training the prediction model. At the end of each data collection trial, the robot opens the gripper and lets the object fall into a recycle box at the bottom of the system. The recycle box will return the object to the same initial position every time so that the robot can automatically start another trial. The reset process is demonstrated in our video supplementary files.


\section{Experiments}
\label{sec:experiments}

In the experimental section, we would like to answer the following questions: (1) How does the prediction model with the learned physical features compare to the one without physical exploration? (2) How does the fusion of the multiple exploration actions compare to each individual action? (3) Can the physical properties of an object be regressed from the learned features? (4) Are objects with similar dynamics close in the embedding space? (5) Can our method accurately swing-up a set of unknown objects to a desired poses consistently?

To answer these questions, we evaluate our method on both \textbf{seen} and \textbf{unseen} objects with a 5-DoF robot arm. Here, ``\textbf{unseen}'' refers to objects with physical proprieties that never appeared in the training set. To assess what information is included in the learned physical feature embedding, we conduct a experiment to directly regress the physical proprieties (friction, mass, center of mass and moment of inertia) from the physical embedding on both \textbf{seen} and \textbf{unseen} objects.

\subsection{Experimental Setup}
\noindent\textbf{Dataset for seen objects:} We collected data with 33 objects with different physical properties as introduced in Sec.~\ref{sec:method}. For the experiments on ``seen'' objects, we split the data of each object into 90\% for training and 10\% for evaluation. Thus, the training set contains 1485 samples (33 objects) and the testing set consists of 165 trials (33 objects).

\noindent\textbf{Dataset for unseen objects:} For the ``unseen'' objects, we split the 33 objects into 27 objects for training and 6 objects (showed in Table.~\ref{tab:exp3}) for testing. The testing set is composed of a combination of 2 different frictions and two different masses placed at 2 different locations. The training set contains 1350 samples and the testing set consists of 300 trials.

\noindent\textbf{Architectures:} We compare five model variants that show case the effectiveness of our design choices:
\begin{itemize}
    \item \textit{None}: No tactile exploration information is given. The model takes the action as input and directly predicts the final swing-up angle.
    \item \textit{PP}: The numerical value of each physical property (friction, mass, center of mass and moment of inertia) of the object is given to the model as inputs.
    \item \textit{Tilting}: The model only process the tactile information of the tilting action into the physical features.
    \item \textit{Shaking}: The model only process the tactile information of the shaking action into the physical features.
    \item \textit{Combined}: Both tactile information of tilting and shaking actions are processed by the fusion model into a joint physical feature embedding.
\end{itemize}

\noindent\textbf{Robot experiment setup:} As shown in Fig.~\ref{fig:setup}, we use a 5-DoF robot arm (ReactorX 150 Robot Arm, Interbotix) for our experiments. For better performance, we replace all the servo motors with DYNAMIXEL XM-430-W350T, ROBOTIS. We use OpenCM9.04 C micro-controller for controlling the robot. In order to get consistent performance, we found that it was critical to send the trajectory to micro-controller buffer in advance and execute on board. Otherwise, the communication latency between PC and micro-controller can produce prohibitively large amounts of actuation noise.

\def\svgwidth{\linewidth}

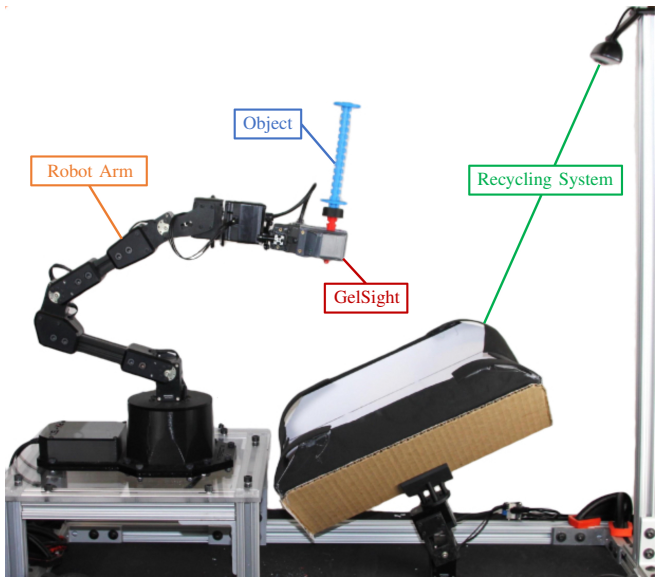
\begin{figure}[t]
	\centering
\begingroup%
  \makeatletter%
  \providecommand\color[2][]{%
    \errmessage{(Inkscape) Color is used for the text in Inkscape, but the package 'color.sty' is not loaded}%
    \renewcommand\color[2][]{}%
  }%
  \providecommand\transparent[1]{%
    \errmessage{(Inkscape) Transparency is used (non-zero) for the text in Inkscape, but the package 'transparent.sty' is not loaded}%
    \renewcommand\transparent[1]{}%
  }%
  \providecommand\rotatebox[2]{#2}%
  \newcommand*\fsize{\dimexpr\f@size pt\relax}%
  \newcommand*\lineheight[1]{\fontsize{\fsize}{#1\fsize}\selectfont}%
  \ifx\svgwidth\undefined%
    \setlength{\unitlength}{615bp}%
    \ifx\svgscale\undefined%
      \relax%
    \else%
      \setlength{\unitlength}{\unitlength * \real{\svgscale}}%
    \fi%
  \else%
    \setlength{\unitlength}{\svgwidth}%
  \fi%
  \global\let\svgwidth\undefined%
  \global\let\svgscale\undefined%
  \makeatother%
  \begin{picture}(1,0.87804878)%
    \lineheight{1}%
    \setlength\tabcolsep{0pt}%
    \put(0,0){\includegraphics[width=\unitlength,page=1]{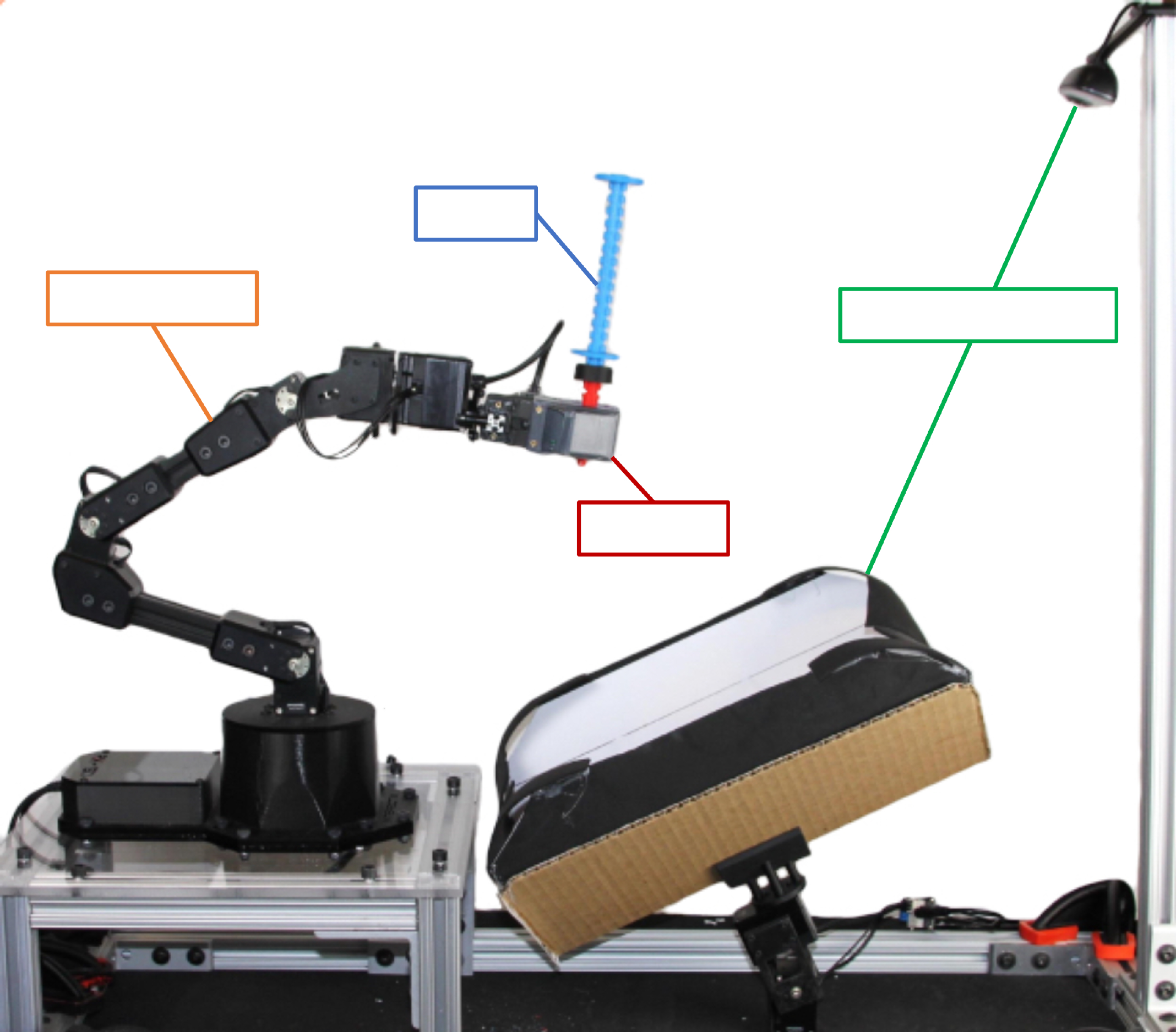}}%
    \put(0.06484823,0.6153784){\color[rgb]{0.92941176,0.49019608,0.19215686}\makebox(0,0)[lt]{\smash{\begin{tabular}[t]{l}\scriptsize Robot Arm\end{tabular}}}}%
    \put(0.36640046,0.68985465){\color[rgb]{0.26666667,0.44705882,0.76862745}\makebox(0,0)[lt]{\smash{\begin{tabular}[t]{l}\scriptsize Object\end{tabular}}}}%
    \put(0.50641071,0.42208036){\color[rgb]{0.75294118,0,0}\makebox(0,0)[lt]{\smash{\begin{tabular}[t]{l}\scriptsize GelSight\end{tabular}}}}%
    \put(0.72688894,0.60351512){\color[rgb]{0,0.69019608,0.31372549}\makebox(0,0)[lt]{\smash{\begin{tabular}[t]{l}\scriptsize Recycling System\end{tabular}}}}%
  \end{picture}%
\endgroup%

    \vspace{-13pt}
	\caption{\textbf{Experiment setup.} The GelSight tactile sensor is mounted on a gripper of the robot arm. The recycling system enables automatic data collection.}
	\vspace{-10pt}
	\label{fig:setup}
\end{figure}

\begin{table}[t]
\begin{tabular}{|l|c|c|c|c|c|c|}
\hline
 & Rand. & None & PP & Tilt. & Shak. & Comb. \\ \hline
Seen & 66.7 & 25.4 & 11.0 & 13.3 & 10.9 & \textbf{10.2} \\ \hline
Unseen & 66.7 & 26.8 & 18.5 & 17.6 & 15.0 & \textbf{12.9} \\ \hline
\end{tabular}
\caption{Quantitative evaluation results of the prediction model with physical embedding from different variants of the fusion model on seen and unseen datasets.}
\label{tab:exp1}
\end{table}

\begin{table*}[t]
\begin{tabular}{|l|c|c|c|c|c|c|c|c|}
\hline
\multirow{2}{*}{} & \multicolumn{4}{c|}{Seen} & \multicolumn{4}{c|}{Unseen} \\ \cline{2-9} 
 & Friction & Mass & Cent. of Mass & Mom. of Iner. & Friction & Mass & Cent. of Mass & Mom. of Iner. \\ \hline
Random & 33.3\% & 0.333 & 0.333 & 0.333 & 33.3\% & 0.333 & 0.333 & 0.333 \\ \hline
Tilting & 89.6\% & 0.101 & 0.150 & 0.090 & 75.6\% & \textbf{0.184} & \textbf{0.086} & 0.141 \\ \hline
Shaking & 96.9\% & 0.121 & 0.203 & 0.184 & 90.1\% & 0.263 & 0.125 & 0.233 \\ \hline
Combined & 94.8\% & 0.085 & 0.135 & 0.112 & \textbf{93.9\%} & 0.200 & 0.099 & \textbf{0.117} \\ \hline
End-to-End & 98.9\% & 0.078 & 0.083 & 0.056 & 95.4\% & 0.073 & 0.110 & 0.095 \\ \hline
\end{tabular}
\caption{Quantitative evaluation results of the physical feature disentanglement on both seen and unseen datasets. The metric for the friction is classification success rate (3 classes). The metric for the rest properties is error in percentage (normalized to 0-1 with the minimum and maximum of the value).}

\vspace{-10pt}
\label{tab:exp2}
\end{table*}


\begin{figure}[t]
	\centering
	\includegraphics[width=1.0\linewidth]{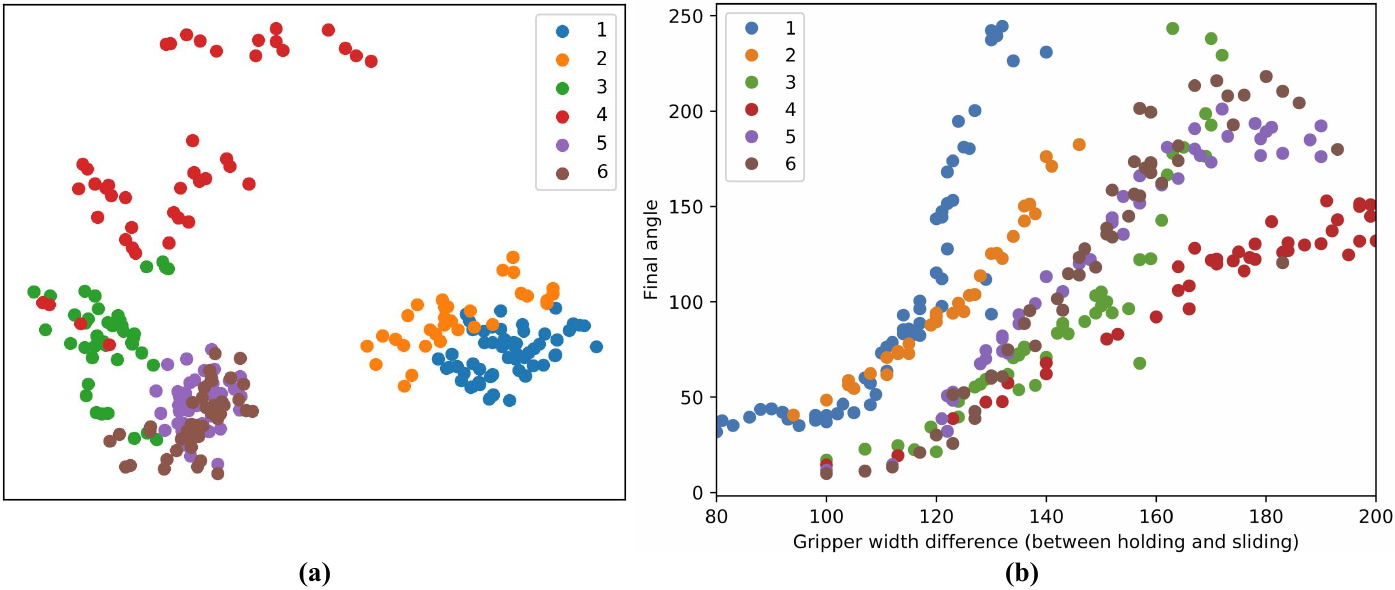}
	\vspace{-13pt}
	\caption{\textbf{Task-oriented physical feature visualization:} (a) Visualization (with PCA) of the outputted physical embedding (\textit{Combined}) on the testing samples of the 6 unseen objects (listed in Table.~\ref{tab:exp3}). (b) Visualization of the data distribution (X-axis: control parameter; Y-axis: final angle) of the testing samples of each object. Each color point refers to one data sample. Objects with similar dynamics are also close to each other on the learned physical embedding space (e.g. 5 and 6). And objects with different dynamics are far away from each other (e.g. 1 and 4).}
	\vspace{-14pt}
	\label{fig:embedding}
\end{figure}

\subsection{Model Performance}
Table.~\ref{tab:exp1} shows the evaluation results of five of our model variants on both seen and unseen datasets. The metric is the error in degrees on the final angle prediction results. Since the baseline method \textit{None} does not have any information about the physical proprieties of the object, it could only output a mean value of the training dataset, showcasing the worst performance. On the other hand, the \textit{Combined} method which uses the fusion model to combine the information from both exploration actions achieves the best results, which surpass the \textit{None} for more than 13\degree on both datasets. This improvement shows the importance of in-hand physical exploration for dynamic manipulation tasks like swing-up.

Also, the \textit{Tilting}, \textit{Shaking} and \textit{Combined} methods outperforms the \textit{PP} baseline method by up to 5 degrees. This is due to the components of the ground truth information being based on the ideal physical model, which has the risk of missing other physical features that also contribute to the model performance such as the elasticity of the contact area of the gripper and the pose of the object in-hand. Since the GelSight tactile sensor provides rich contact information on the finger tip, methods relying on this information have the potential to learn their own joint physical understanding about the held object and the system. These results showcase the advantage of using an intuitive physics reasoning compared to manually engineering physical features.

In the ablation study between \textit{Tilting}, \textit{Shaking} and \textit{Combined}, the performance of the methods with individual exploration action is inferior to the combined version, especially for the unseen situation. Hence, we conducted an additional experiment to evaluate what information is learned in each exploration action and why the combined version could achieve the best performance.

\subsection{Physical Feature Disentanglement}
We use a three-layer MLP as a disentangle network which takes the physical feature embedding as input and regresses to the numerical values of mass, center of mass and moment of inertia. Another branch of the network outputs a classification result for the friction (3 classes). The training and testing data for both seen and unseen situations follow the same setting as the prediction model. The weights of the network that generates the physical embedding are fixed and only the disentangle network is trained and tested. In addition to the model variants introduced previously, we add another \textit{End-to-End} method which trains the whole pipeline to output the physical properties. This method can be regarded as the best performance that the model can reach.

Table.~\ref{tab:exp2} shows the experimental results of the physical feature disentanglement. The metric of the friction is the classification success rate. The metric for the rest of the physical properties is the error in percentage, where each property is normalized to 0-1 based on the minimum and maximum value. For both seen and unseen situation, all the model variants outperform the \textit{Random} baseline, which proves that all of the physical properties are included in the learned embedding.

In addition, we can observe from the results the difference in focus between each of the exploration actions. For instance, the \textit{tilting} action is good at reasoning the mass and center of mass, which surpasses the \textit{shaking} for 8\% in mass and 4\% in center of mass on the unseen situation. This is mainly because the tilting action provides stable torque force signal by placing the held object at different angles, for which is easier to calculate these properties compared to \textit{shaking}. On the other hand, the \textit{shaking} action achieves 93.9\% friction classification success rate which is higher than \textit{tilting} action by 15\%. This is to the fact that, as opposed to \textit{tilting} when the object is held firmly, \textit{shaking} loosens the gripper to enable in-hand sliding of the object, capturing friction information. Because of the loosening if the gripper, \textit{shaking} failed to acquire the mass and center of mass information, which requires stable observations. It is surprising to find that the moment of inertia of the \textit{tilting} also outperforms the \textit{shaking}. One of the possible reasons for this is the model inferring the moment of inertia based on its understanding of mass and center of mass. The \textit{combined} method successfully fuses the information from both actions and achieved a balanced performance among all the physical properties. This experiments show the importance of fusing multiple exploration actions and why the \textit{combined} method could reach the best prediction results.


\subsection{Task-oriented Physical Feature}
Another advantage of learning joint physical features compared to estimating each property individually is its potential to generate task-oriented feature embeddings, where the objects close to each other in the embedding space can share similar control policies. We use PCA~\cite{wold1987principal} to project the learned physical embeddings from the \textit{Combined} method to points on a 2D plot and visualize all the testing results of the 6 unseen testing objects in Fig.~\ref{fig:embedding}(a). We also visualize the data distribution of these test samples in Fig.~\ref{fig:embedding}(b), where the X-axis refers to the control parameter and the Y-axis is the final swing-up angle. As we can see, for objects with similar policy distribution (objects 5 and 6), the distance between their embedding is also short. And for objects with large differences in the policy distribution (objects 1 and 4), the distance between their embedding is large. This result confirms that the learned physical embeddings are indeed task-oriented, which largely benefits the dynamic swing-up manipulation.

\subsection{Swing-up Results}

\begin{table}[t]
\centering
\begin{tabular}{|c|c|c|c|c|c|}
\hline
ID & Objects & Errors & ID & Objects & Errors\\ \hline
 
1 & \includegraphics[width=0.18\linewidth]{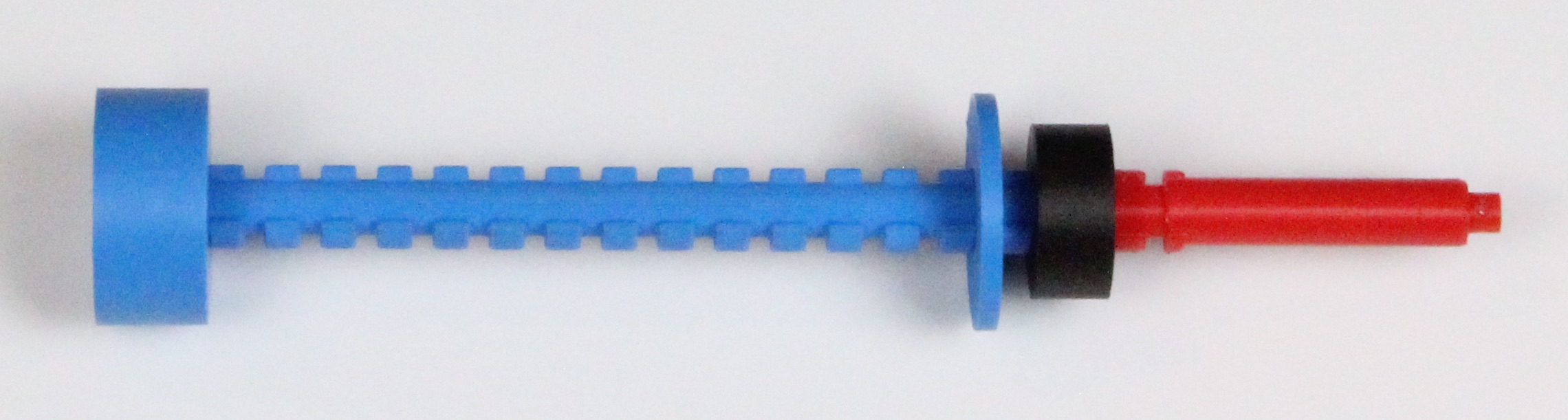} & 21.4 &

2 & \includegraphics[width=0.18\linewidth]{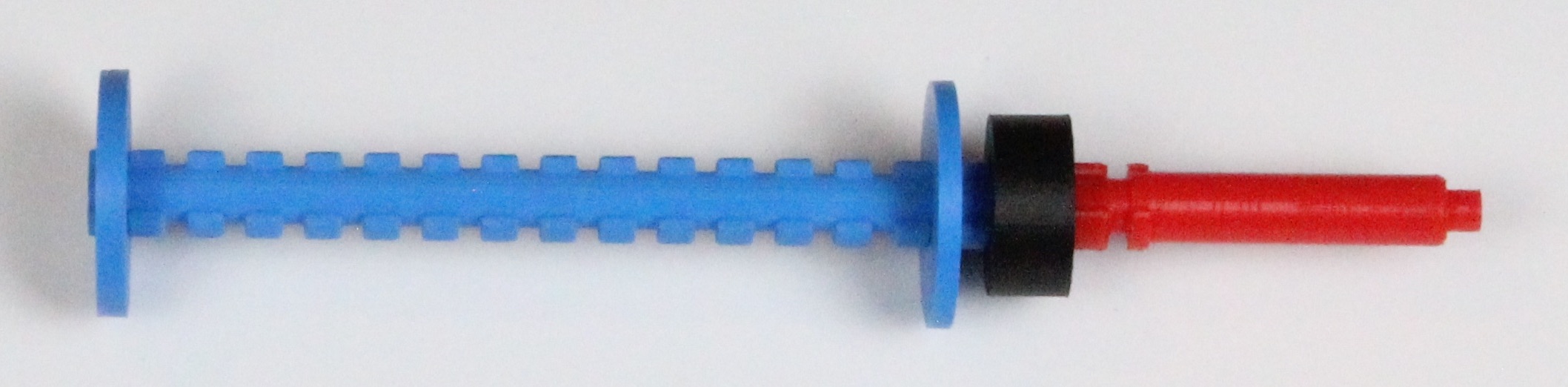} &12.3\\ \hline

3 & \includegraphics[width=0.18\linewidth]{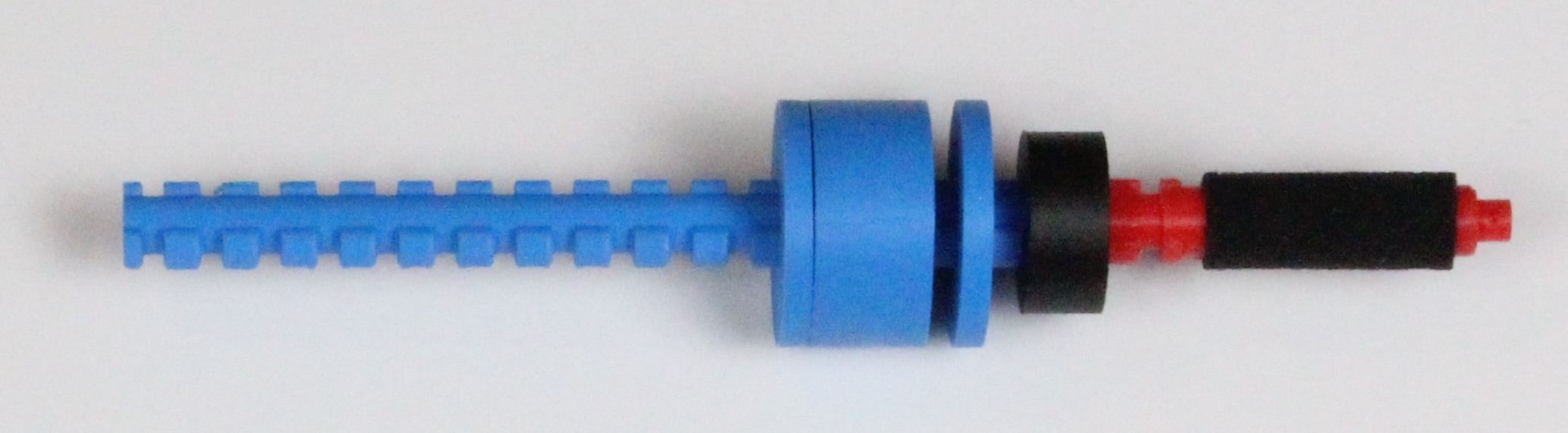} & 19.8 &

4 & \includegraphics[width=0.18\linewidth]{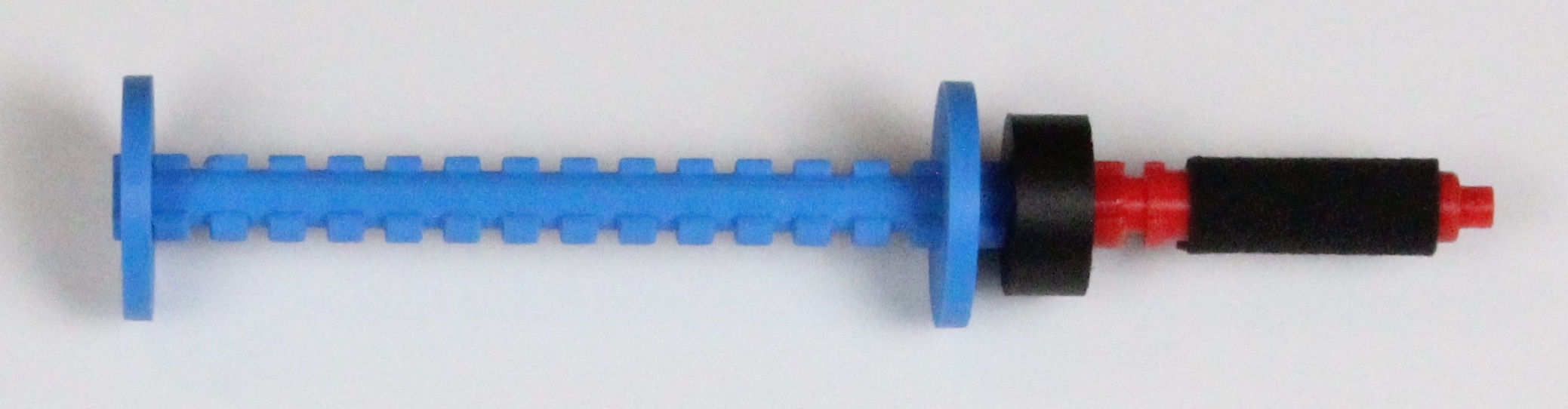} & 8.3\\ \hline

5 & \includegraphics[width=0.18\linewidth]{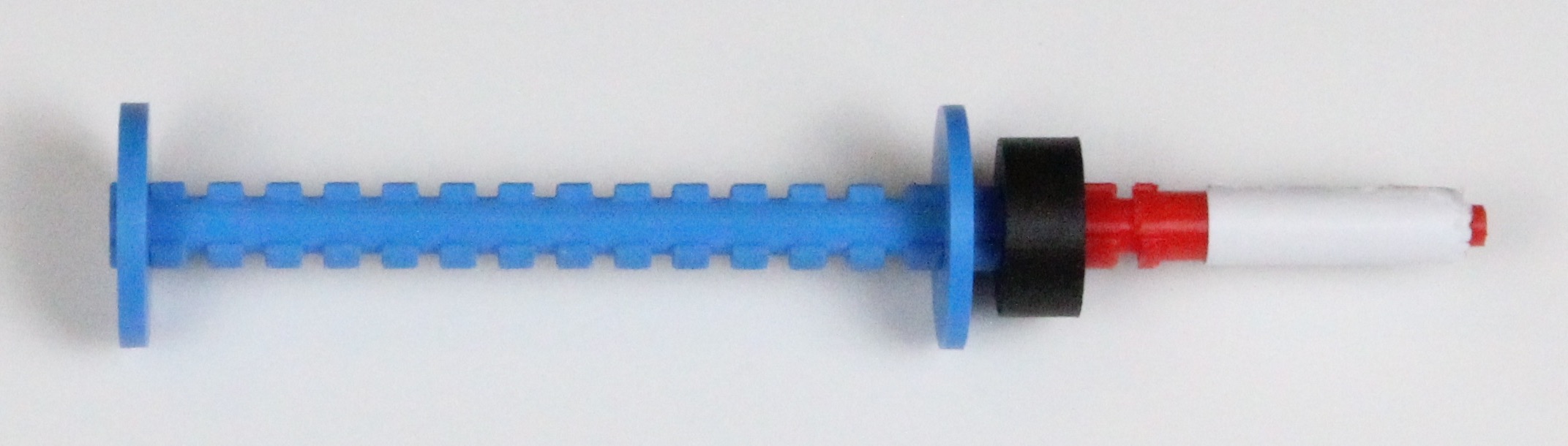} & 18.3 &

6 & \includegraphics[width=0.18\linewidth]{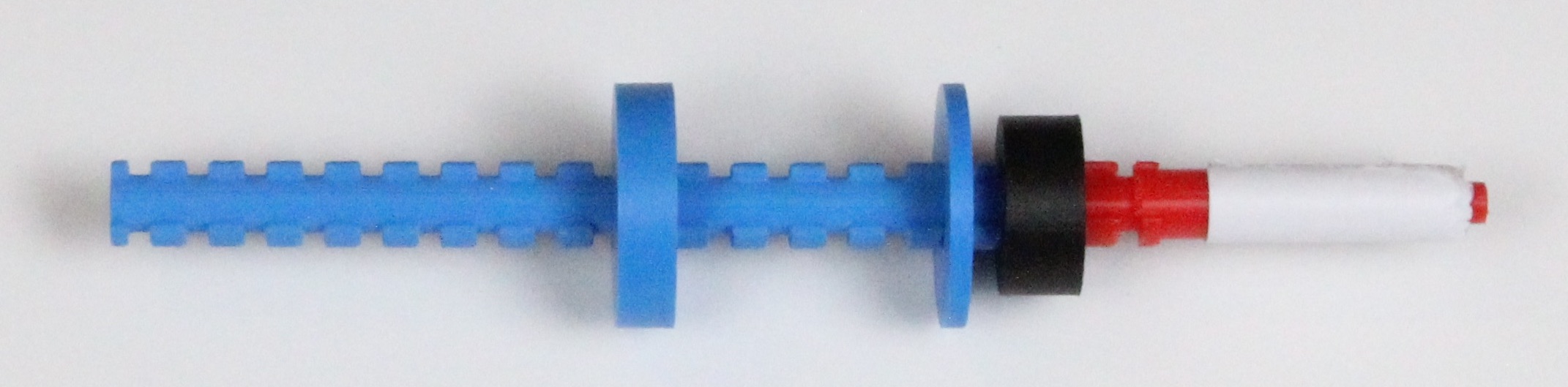} & 23.4\\ \hline

Mean & \multicolumn{5}{c|}{17.2}  \\ \hline

\end{tabular}
\caption{Swing-up results on 6 unseen testing objects (with ID 1-6 same as Fig.~\ref{fig:embedding}). The robot uniformly samples a set of actions and selects the one with the prediction result closest to the final goal to perform the task. In this table, each object is tested 20 trials (5 trials for each desired angle: 45\degree, 90\degree, 135\degree and 180\degree) and the mean error is listed.}

\vspace{-10pt}
\label{tab:exp3}
\end{table}

We deploy the trained model of the \textit{Combine} method in the robot arm. Given the target angle, the robot samples different control parameters, and chooses the one whose prediction is closest to the target angle. We test 20 times (5 times each for 45, 90, 135, 180 degrees as target angle) for each unseen object. The robot is able to adapt the control policy automatically for objects with different physical properties. The evaluation shows that the model performs better on lighter objects which have less uncertainty compared to heavier objects. The detailed results are shown in Table.~\ref{tab:exp3}.  

\section{Discussion and Future Work}
\label{sec:conclusion}

We have presented SwingBot, a robot system that identifies physical features of held objects from tactile exploration, providing crucial information for a dynamic swing-up manipulation. SwingBot is based on a novel multi-action fusion network that combines the information acquired via multiple exploration actions into a joint embedding space. The whole pipeline is trained in an end-to-end self-supervised manner. We used the performance of the swing-up task to compare our embedding with variants trained with single actions and with swing-up actions that do not consider any form of tactile information. These comparisons showed that swing-up actions that relied on our fusion method achieved the best performances. Furthermore, we showed that the learned task-oriented feature embedding could also be used to successfully regress individual physical properties such as mass, center of mass, moment of inertia and friction. 

Current limitations are inherently coupled to the fact that our analysis of the embedding is based on the performance of a single task. This task is very specific and heavily conditioned by the available hardware. The robot platform that was used suffers from high actuation noise increasing the error of the swing-up angle predictions.

In addition, while the GelSight sensors provide very rich information, the current sensing latencies prevent the observation of the full swing-up movement. Using a more robust robotic system in conjunction with a GelSight sensor with lower latencies would potentially enable the use of real time feedback control as opposed to the open loop solution that was proposed. 

Regarding future work, one interesting direction is to learn the optimal exploration actions by using the quality of the resulting embeddings to guide the learning. Another interesting direction is to assess how useful these embeddings are for other task and if an embedding learn for one task can be transferred onto other tasks.




\section*{Acknowledgments}

\addtolength{\textheight}{-0cm}   

Toyota Research Institute (TRI), and the Office of Naval Research (ONR) [N00014-18-1-2815] provided funds to support this work.

\bibliographystyle{IEEEtran}
\bibliography{references}

\end{document}